\documentclass[11pt,letterpaper]{article}
\usepackage{ijcnlp2017}
\usepackage{times}
\usepackage{latexsym}
\usepackage{graphicx}
                                                     
\usepackage{url}
\usepackage{bm}
\usepackage{float}
\usepackage{multirow}
\usepackage{amsmath, amssymb}
\ijcnlpfinalcopy

\title{Local Monotonic Attention Mechanism \\for End-to-End Speech and Language Processing}

\author{Andros Tjandra, Sakriani Sakti, \and Satoshi Nakamura \\
	Graduate School of Information Science \\
	Nara Institute of Science and Technology, Japan\\
	{\tt \{andros.tjandra.ai6, ssakti, s-nakamura\}@is.naist.jp}}

\date{}

\begin{document}

\maketitle

\begin{abstract}
	Recently, encoder-decoder neural networks have shown impressive performance on many sequence-related tasks. The architecture commonly uses an attentional mechanism which allows the model to learn alignments between the source and the target sequence. Most attentional mechanisms used today is based on a global attention property which requires a computation of a weighted summarization of the whole input sequence generated by encoder states. However, it is computationally expensive and often produces misalignment on the longer input sequence. Furthermore, it does not fit with monotonous or left-to-right nature in several tasks, such as automatic speech recognition (ASR), grapheme-to-phoneme (G2P), etc. In this paper, we propose a novel attention mechanism that has local and monotonic properties. Various  ways  to  control  those  properties  are  also  explored. Experimental results on ASR, G2P and machine translation between two languages with similar sentence structures, demonstrate that
	the proposed encoder-decoder model with local monotonic attention could achieve significant performance improvements and reduce the computational complexity in comparison with the one that used the standard global attention architecture.
\end{abstract}

\section{Introduction}

End-to-end training is a newly emerging approach to sequence-to-sequence mapping tasks, that allows the model to directly learn the
mapping between variable-length representation of different modalities (i.e., text-to-text sequence \cite{bahdanau2014neural, sutskever2014sequence}, speech-to-text sequence \cite{chorowski2014end, chan2016listen}, image-to-text
sequence \cite{kelvinxu2016image2caption}, etc). 

%In the context of a speech recognition task, it replaces the conventional fashion that consists of several sub-components that are trained and tuned separately, which include a front-end feature extractor, a acoustic modeling, a pronunciation lexicon, and a language model, into a single integrated neural network that is jointly optimized at the same time to directly map from a speech sequence to a text transcription. Besides speech recognition, sequence-to-sequence task can be applied to language and text related problems, for example: simplifying the conventional phrase-based machine translation (MT) by integrate multiple modules (e.g word alignment, phrase pair extraction, lexical reordering, etc) into a single end-to-end model.

One popular approaches in the end-to-end mapping tasks of different modalities is based on encoder-decoder architecture.
The earlier version of an encoder-decoder model is built with only two different components \cite{sutskever2014sequence, cho2014learning}: (1) an encoder that processes the source sequence and encodes them into a fixed-length vector; and (2) a decoder that produces the target sequence based on information from fixed-length vector given by encoder. Both the encoder and decoder are jointly trained to maximize the probability of
a correct target sequence given a source sequence. This architecture has been applied in many applications such as machine translation \cite{sutskever2014sequence, cho2014learning}, image captioning \cite{karpathy2015deep}, and so on.

However, such architecture encounters difficulties, especially for coping with long sequences. Because in order to generate the correct target sequence, the decoder solely depends only on the last hidden state of the encoder. In other words, the network needs to compress all of the information contained in the source sequence into a single fixed-length vector. \cite{cho2014properties} demonstrated a decrease in the performance of the encoder-decoder model associated with an increase in the length of the input sentence sequence. Therefore, \cite{bahdanau2014neural} introduced attention mechanism to address these issues. Instead of relying on a fixed-length vector, the decoder is assisted by the attention module to get the related context from the encoder sides, depends on the current decoder states.

Most attention-based encoder-decoder model used today has a ``global" property \cite{bahdanau2014neural, luong2015effective}. Every time the decoder needs to predict the output given the previous output, it must compute a weighted summarization of the whole input sequence generated by the encoder states. This global property allows the decoder to address any parts of the source sequence at each step of the output generation and provides advantages in some cases like machine translation tasks. Specifically, when the source and the target languages have different sentence structures and the last part of the target sequence may depend on the first part of the source sequence. However, although the global attention mechanism has often improved performance in some tasks, it is very computationally expensive. For a case that requires mapping between long sequences, misalignments might happen in standard attention mechanism  \cite{kim2017joint}. Furthermore, it does not fit with monotonous or left-to-right natures in several tasks, such as ASR, G2P, etc.

In this paper, we propose a novel attention module that has two important characteristics to address those problems: local and monotonicity properties. The local property helps our attention module focus on certain parts from the source sequence that the decoder wants to transcribe, and the monotonicity property strictly generates alignment left-to-right from beginning to the end of the source sequence. In case of speech recognition task that need to produces a transcription given the speech signal, the attention module is now able to focus on the audio's specific timing and always move in one direction from the start to the end of the audio. Similar way can be applied also for G2P or machine translation (MT) between two languages with similar sentences structure, i.e., Subject-Verb-Object (SVO) word order in English and French languages. Experimental results demonstrate that the proposed encoder-decoder model with local monotonic attention could achieve significant performance improvements and reduce the computational complexity in comparison with the one that used the standard global attention architecture.

\section{Attention-based Encoder Decoder Neural Network}
\begin{figure}[]
	\centering
	\includegraphics[width=0.85\linewidth]{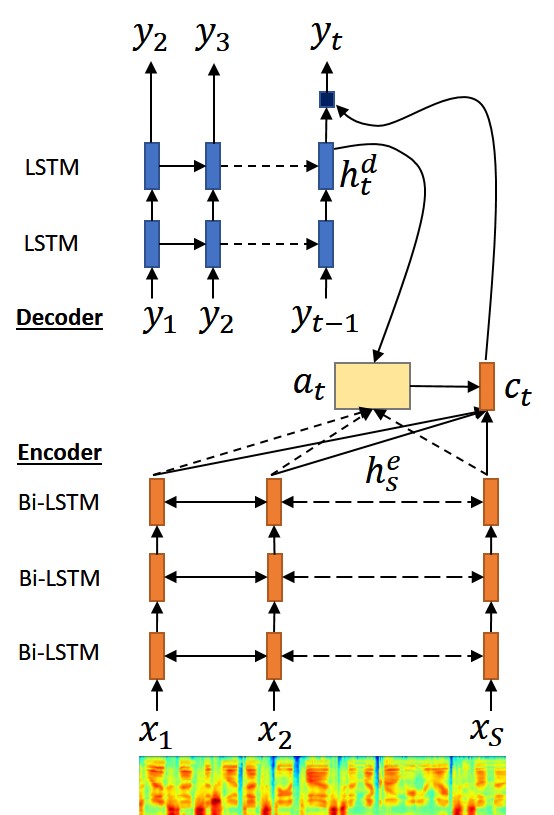}
	\caption{Attention-based encoder-decoder architecture.}
	\label{fig:atte2e}
\end{figure}

The encoder-decoder model is a neural network that directly models conditional probability $p(\mathbf{y}|\mathbf{x})$, where $\mathbf{x} = [x_1, ..., x_S]$ is the source sequence with length $S$ and $\mathbf{y} = [y_1, ..., y_T]$ is the target sequence with length $T$. Figure~\ref{fig:atte2e} shows the overall structure of the attention-based encoder-decoder model that consists of encoder, decoder and attention modules.

The encoder task processes input sequence $\mathbf{x}$ and outputs representative information $\mathbf{h^e} = [h^e_1, ...,h^e_S]$ for the decoder. The attention module is an extension scheme for assisting the decoder to find relevant information on the encoder side based on the current decoder hidden states \cite{bahdanau2014neural, luong2015effective}. Usually, attention modules produces context information $c_t$ at the time $t$ based on the encoder and decoder hidden states:
\begin{align}
c_t &= \sum_{s=1}^{S} a_t(s) * h^e_s \\
a_t(s) &= \text{Align}({h^e_s}, h^d_t) \nonumber \label{eq:align} \\
&= \frac{\exp(\text{Score}(h^e_s, h^d_t))}{\sum_{s=1}^{S}\exp(\text{Score}(h^e_s, h^d_t))}
\end{align}
There are several variations for score functions:
\begin{align}
\text{Score}(h_s^e, h_t^d) =
\begin{cases}
\langle h_s^e, h_t^d\rangle, & \text{dot product}  \\
h_s^{e\intercal} W_{s} h_t^d, & \text{bilinear}  \\
V_s^{\intercal} \tanh(W_{s} [h_s^e, h_t^d]), & \text{MLP} \label{eq:mlpscore}  \\
\end{cases}
\end{align} where $\text{Score}:(\mathbb{R}^M \times \mathbb{R}^N) \rightarrow \mathbb{R}$, $M$ is the number of hidden units for encoder and $N$ is the number of hidden units for decoder.
Finally, the decoder task, which predicts the target sequence probability at time $t$ based on previous output and context information $c_t$ can be formulated:
\begin{equation}
\log{p(\mathbf{y}|\mathbf{x})} = \sum_{t=1}^{T}\log{p(y_t|y_{<t}, c_t)}
\end{equation}
For speech recognition task, most common input $\mathbf{x}$ is a sequence of feature vectors like Mel-spectral filterbank and/or MFCC. Therefore, $\mathbf{x} \in \mathbb{R}^{S \times D}$ where D is the number of features and S is the total frame length for an utterance. Output $\mathbf{y}$, which is a speech transcription sequence, can be either phoneme or grapheme (character) sequence.
In text-related task such as machine translation, $\textbf{x}$ and $\textbf{y}$ are a sequence of word or character indexes.

\section{Locality and Monotonicity Properties}
In the previous section, we explained the standard global attention-based encoder-decoder model. However, in order to control the area and focus attention given previous information, such mechanism requires to apply the scoring function into all the encoder states and normalizes them with a softmax function. Another problem is we cannot explicitly enforce the probability mass generated by the current attention modules that are always moving incrementally to the end of the source sequence. In this section, we discuss and explain how to model the locality and monotonicity properties on the attention module. This way, we could improve the sensitivity of capturing regularities and ensure to focus only an important subset instead of whole sequence.

\begin{figure}[]
	\centering
	\includegraphics[width=0.7\linewidth]{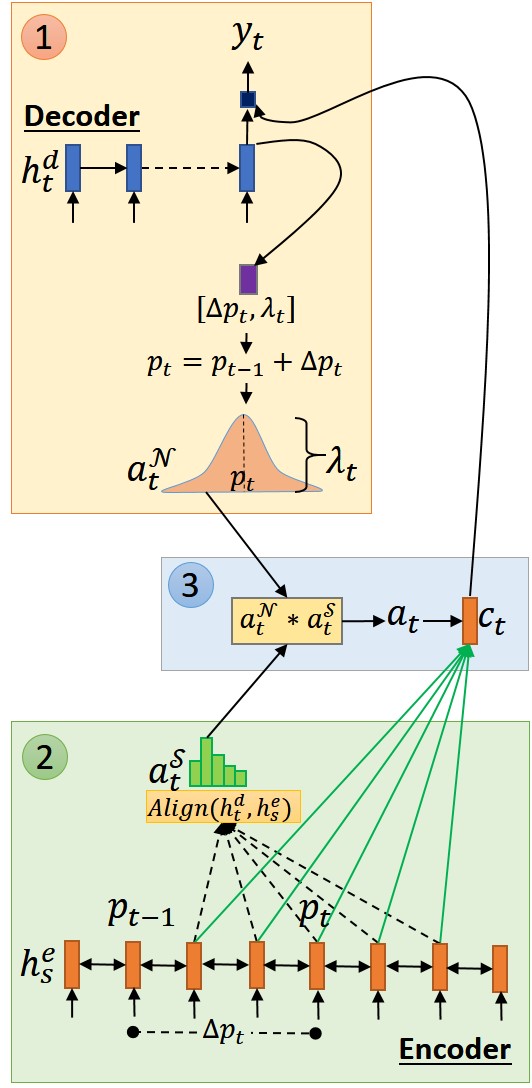}
	\caption{Local monotonic attention.}
	\label{fig:localmono}
\end{figure}

Figure~\ref{fig:localmono} illustrates the overall mechanism of our proposed local monotonic attention, and details are described blow.
\begin{enumerate}
	\item \textbf{Monotonicity-based Prediction of Central Position}\\
	First, we define how to predict the next central position of the alignment illustrated in Part (1) of Figure~\ref{fig:localmono}. Assume we have source sequence with length $S$, which is encoded by the stack of Bi-LSTM (see Figure~\ref{fig:atte2e}) into $S$ encoded states $\mathbf{h^e} = [h_1^e, ..., h_S^e]$. At time $t$, we want to decode the $t$-th target output given the source sequence, previous output $y_{t-1}$, and current decoder hidden states $h_t^d \in \mathbb{R}^{N}$. In standard approaches, we use hidden states $h_t^d$ to predict the position difference $\Delta p_{t}$ with a multilayer perceptron (MLP). We use variable $\Delta p_t$ to determine how far we should move the center of the alignment compared to previous center $p_{t-1}$.
	
	In this paper, we propose two different formulations for estimating $\Delta p_t$ to ensure a forward or monotonicity movement:
	\begin{itemize}
		\item \textbf{Constrained position prediction:}\\
		We limit maximum range from $\Delta p_t$ with hyperparameter $C_{max}$ with the following equation:
		\begin{align} \label{eq:constrained_pos}
		\Delta p_{t} = C_{max} * \text{sigmoid}(V_p^\intercal \tanh(W_p h^d_t))
		\end{align}
		Here we can control how far our next center of alignment position $p_t$ relies on our datasets and guarantee $0 \le \Delta p_t \le C_{max}$. However, it requires us to handle hyperparameter $C_{max}$.\\
		
		\item \textbf{Unconstrained position prediction:}\\
		Compared to a previous formulation, since we do not limit the maximum range of $\Delta p_t$, here we can ignore hyperparameter $C_{max}$ and use exponential ($\exp$) function instead of sigmoid. We can also use another function (e.g softplus) as long as the function satisfy $f : \mathbb{R} \rightarrow \mathbb{R}_{0}^{+}$ and the result of $\Delta p_{t} \geq 0$. We formulate unconstrained position prediction with following equation:
		\begin{align}
		\Delta p_t = \exp(V_p^\intercal \tanh(W_p h^d_t))
		\end{align}
	\end{itemize}
	
	Here $V_p \in \mathbb{R}^{K \times 1}$, $W_p \in \mathbb{R}^{K \times N}$, $N$ is the number of decoder hidden units and $K$ is the number of hidden projection layer units. We omit the bias for simplicity. Both equations guarantee monotonicity properties since $\forall t \in [1..T], p_t \geq (p_{t-1} + \Delta p_t$).
	
	Additionally, we also used scaling variable $\lambda_t$ to scale the unnormalized Gaussian distribution that depends on $h_t$. We calculated $\lambda_t$ with following equation:
	\begin{align}
	\lambda_t = \exp(V_{\lambda}^\intercal \tanh(W_p h^d_t))
	\end{align} where $V_\lambda \in \mathbb{R}^{K \times 1}$. In our initial experiments, we discovered that we improved our model performance by scaling with $\lambda_t$ for each time-step. The main objective of this step is to generate a scaled Gaussian distribution $a_t^\mathcal{N}$:
	\begin{align}
	a_t^\mathcal{N}(s) = \lambda_t * \exp\left(-\frac{(s-p_t)^2}{2\sigma^2}\right).
	\end{align}  where $p_t$ is the mean and $\sigma$ is the standard deviation, both of which are used to calculate the weighted sum from the encoder states to generate context vector $c_t$ later. In this paper, we treat $\sigma$ as a hyperparameter.
	
	\item \textbf{Locality-based Alignment Generation}\\
	After calculating new position $p_t$, we generate locality-based alignment, as shown in Part (2) of Figure~\ref{fig:localmono}. Based on predicted position $p_t$, we follow \cite{luong2015effective} to generate alignment $a_t^\mathcal{S}$ only within $[p_t-2\sigma, p_t+2\sigma]$:
	\begin{eqnarray}
	a_t^\mathcal{S}(s) = \text{Align}(h_s^e, h_t^d), \\ \nonumber
	\forall s\in[p_t-2\sigma, p_t+2\sigma].
	\end{eqnarray}
	Since $p_t$ is a real number and the indexes for the encoder states are integers, we convert $p_t$ into an integer with floor operation. 
	After we know the center of the position $p_t$, we only need to calculate the scores (Eq.~\ref{eq:mlpscore}) for each encoder states in $[p_t-2\sigma, .., p_t+2\sigma]$ then calculate the context alignment scores (Eq.~\ref{eq:align}).
	
	Compared to the standard global attention, we can reduce the decoding computational complexity $O(T*S)$ into $O(T*\sigma)$ where $\sigma \ll S$ and $\sigma$ is constant, $T$ is total decoding step, $S$ is the length of the encoder states. 
	
	\item \textbf{Context Calculation}\\
	In the last step, we calculate context $c_t$ with alignments $a_t^\mathcal{N}$ and $a_t^\mathcal{S}$, as shown in Part (3) of Figure~\ref{fig:localmono}:
	\begin{align}
	c_t = \sum_{s=(p_t-2\sigma)}^{(p_t+2\sigma)} \left(a_t^\mathcal{N}(s) * a_t^\mathcal{S}(s)\right) * h_s^e
	\end{align}
	Context $c_t$ and current hidden state $h_t^d$ will later be utilized for calculating current output $y_t$.
\end{enumerate}

Overall, we can rephrase the first step as generating ``prior" probabilities $a_t^\mathcal{N}$ based on the previous $p_{t-1}$ position and the current decoder states. Then the second step task generates ``likelihood" probabilities $a_t^\mathcal{S}$ by measuring the relevance of our encoder states with the current decoder states. In the third step, we combine our ``prior" and ``likelihood" probability into an unnormalized ``posterior" probability $a_t$ and calculate expected context $c_t$.

\section{Experiment on Speech Recognition}

We applied our proposed architecture on ASR task. The local property helps our attention module focus on certain parts from the speech that the decoder wants to transcribe, and the monotonicity property strictly generates alignment left-to-right from beginning to the end of the speech.

\subsection{Speech Data}

We conducted our experiments on the TIMIT \footnote{\url{https://catalog.ldc.upenn.edu/ldc93s1}} \cite{garofolo1993darpa} dataset with the same set-up for training, development, and test sets as defined in the Kaldi s5 recipe \cite{povey11asru}. The training set contains 3696 sentences from 462 speakers. We also used another sets of 50 speakers for the development set and the test set contains 192 utterances, 8 each from 24 speakers. For every experiment, we used 40-dimensional fbank with delta and acceleration (total 120-dimension feature vector) extracted from the Kaldi toolkit. The input features were normalized by subtracting the mean and divided by the standard deviation from the training set. For our decoder target, we re-mapped the original target phoneme set from 61 into 39 phoneme class plus the end of sequence mark (eos).
%https://perso.limsi.fr/lamel/TIMIT_NISTIR4930.pdf

\subsection{Model Architectures}

On the encoder sides, we projected our input features with a linear layer with 512 hidden units followed by tanh activation function. We used three bidirectional LSTMs (Bi-LSTM) for our encoder with 256 hidden units for each LSTM (total 512 hidden units for Bi-LSTM). To reduce the computational time, we used hierarchical subsampling \cite{graves2012supervised, bahdanau2016end}, applied it to the top two Bi-LSTM layers, and reduced their length by a factor of 4.

On the decoder sides, we used a 64-dimensional embedding matrix to transform the input phonemes into a continuous vector, followed by two unidirectional LSTMs with 512 hidden units. For every local monotonic model, we used an MLP with 256 hidden units to generate $\Delta p_t$ and $\lambda_t$. Hyperparameter $2\sigma$ was set to $3$, and $C_{max}$ for constrained position prediction (see Eq. \ref{eq:constrained_pos}) was set to $5$. Both hyperparameters were empirically selected and generally gave consistent results across various settings in our proposed model. For our scorer module, we used bilinear and MLP scorers (see Eq~ \ref{eq:mlpscore}) with 256 hidden units. We used an Adam \cite{kingma2014adam} optimizer with a learning rate of $5e-4$.

In the recognition phase, we generated transcriptions with best-1 (greedy) search from the decoder. We did not use any language model in this work. All of our models were implemented on the Chainer framework \cite{chainer_learningsys2015}.

For comparison, we evaluated our proposed model with the standard global attention-based encoder-decoder model and \textit{local-m} attention \cite{luong2015effective} as the baseline. Most of the configurations follow the above descriptions, except the baseline model that does not have an MLP for generating $\Delta p_t$ and $\lambda_t$.

\section{Result and Discussion for Speech Recognition}

Table \ref{tbl:timit} summarizes our experiments on our proposed local attention models and compares them to the baseline model using several possible scenarios.

\subsection{Constrained vs Unconstrained Position Prediction}

Considering the use of constrained and unconstrained position prediction $\Delta p_t$, our results show that the model with the unconstrained position prediction ($\exp$) model gives better results than one based on the constrained position prediction (sigmoid) model on both MLP and bilinear scorers. We conclude that it is more beneficial to use the unconstrained position prediction formulation since it gives better performance and we do not need to handle the additional hyperparameter $C_{max}$.

\subsection{Alignment Scorer vs Non-Scorer}

Next we investigate the importance of the scorer module by comparing the results between a model with and without it.
Our results reveal that, by only relying on Gaussian alignment $a_t^{\mathcal{N}}$ and set $a_t^{\mathcal{S}} = 1$, our model performance's was worse than one that used both the scorer and Gaussian alignment. This might be because the scorer modules are able to correct the details from the Gaussian alignment based on the relevance of the encoder states in the current decoder states. Thus, we conclude that alignment with the scorer is essential for our proposed models.

\begin{table}[]
	\centering
	\footnotesize
	\caption{Results from baseline and proposed models on ASR task with TIMIT test set.}
	\vspace{0.2cm}
	\label{tbl:timit}
	\begin{tabular}{|l|c|l|c|}
		\hline
		\multicolumn{3}{|c|}{\textbf{Model}}                                                                                                                                                                                                       & \textbf{\begin{tabular}[c]{@{}c@{}}Test \\ PER (\%)\end{tabular}} \\ \hline
		\hline
		\multicolumn{4}{|c|}{\textbf{Global Attention Model (Baseline)}}                                                                                                                                                                                                    \\ \hline
		\multicolumn{3}{|l|}{\begin{tabular}[c]{@{}c@{}}Att Enc-Dec (pretrained with HMM align)\\ \cite{chorowski2014end}\end{tabular}}
		& 18.6                 \\ \hline
		\multicolumn{3}{|l|}{Att Enc-Dec \cite{pereyra2017regularizing}}                                                                                                                                                                         & 23.2                   \\ \hline
		\multicolumn{3}{|l|}{Att Enc-Dec \cite{luo2016learning}}                                                                                                                                                                                 & 24.5                   \\ \hline
		\multicolumn{3}{|l|}{Att Enc-Dec with MLP Scorer (ours)}                                                                                                                                                                                   & 23.8                   \\ \hline
		\multicolumn{3}{|l|}{
			\begin{tabular}[c]{@{}c@{}}Att Enc-Dec with \textit{local-m} (ours)\\ \cite{luong2015effective}     \end{tabular}} & - \\ \hline
		\hline
		\multicolumn{4}{|c|}{\textbf{Local Attention Model (Proposed)}}                                                                                                                                                                                                     \\ \hline
		\multicolumn{1}{|c|}{\textbf{Monotonicity}} & \multicolumn{2}{c|}{\textbf{Locality}} & \multicolumn{1}{c|}{}\\ \hline
		\textbf{\begin{tabular}[c]{@{}c@{}}Pos Prediction\\ $\Delta p_{t}$\end{tabular}}
		& \textbf{\begin{tabular}[c]{@{}c@{}}Alignment\\ $\text{Score}(h_s^e, h_t^d)$\end{tabular}}
		& \multicolumn{1}{c|}{\textbf{\begin{tabular}[c]{@{}c@{}}Func. \\ Type\end{tabular}}}
		& \textbf{\begin{tabular}[c]{@{}c@{}}Test \\ PER (\%)\end{tabular}} \\ \hline
		Const ($sigmoid$)    & No      & -         & 23.2                   \\ \hline
		Const ($sigmoid$)    & Yes     & Bilinear  & 21.9                   \\ \hline
		Const ($sigmoid$)    & Yes     & MLP       & 21.7                   \\ \hline
		\hline
		Unconst ($exp$)   & No      & -         & 23.1                     \\ \hline
		Unconst ($exp$)   & Yes     & Bilinear  & \textbf{20.9}                   \\ \hline
		Unconst ($exp$)   & Yes     & MLP       & 21.4                   \\ \hline
	\end{tabular}
\end{table}

\subsection{Overall comparison to the baseline}

Overall, our proposed encoder-decoder model with local monotonic attention significantly improved the performance and reduced the computational complexity in comparison with one that used standard global attention mechanism (we cannot compare directly with \cite{chorowski2014end} since its pretrained with HMM state alignment). We also tried \textit{local-m} attention from \cite{luong2015effective}, however our model cannot converge and we hypothesize the reason is because ratio length between the speech and their corresponding text is larger than 1, therefore the $\Delta p_t$ cannot be represented by fixed value. The best performance achieved by our proposed model with unconstrained position prediction and bilinear scorer, and provided 12.2\% relative error rate reduction to our baseline.

\section{Experiment on Grapheme-to-Phoneme}

We also investigated our proposed architecture on G2P conversion task. Here, the model need to generate corresponding phoneme given small segment of characters and its always moving from left to right. The local property helps our attention module focus on certain parts from the grapheme source sequence that the decoder wants to convert into phoneme, and the monotonicity property strictly generates alignment left-to-right from beginning to the end of the grapheme source sequence.

\subsection{Dataset}
Here, we used the CMUDict dataset\footnote{CMUdict: \url{https://sourceforge.net/projects/cmusphinx/files/G2P\%20Models/phonetisaurus-cmudict-split.tar.gz}}. It contains 113438 words for training and 12753 for testing (12000 unique words). For validation, we randomly select 3000 sentences from the training set. The evaluation metrics for this task are phoneme error rate (PER) and word error rate (WER). In the evaluation process, there are some words has multiple references (pronunciations). Therefore, we select one of the references that has lowest PER between compared to our hypothesis, and if the hypothesis completely match with one of those references, then the WER is not increasing. For our encoder input, we used 26 letter (A-Z) + single quotes ('). For our decoder target, we used 39 phonemes plus the end of sequence mark (eos).

\subsection{Model Architectures}
On the encoder sides, the characters input were projected into 256 dims using embedding matrix. We used two bidirectional LSTMs (Bi-LSTM) for our encoder with 512 hidden units for each LSTM (total 1024 hidden units for Bi-LSTM). On the decoder sides, the phonemes input were projected into 256 dims using embedding matrix, followed by two unidirectional LSTMs with 512 hidden units. For local monotonic model, we used an MLP with 256 hidden units to generate $\Delta p_t$ and $\lambda_t$. For this task, we only used the unconstrained formulation because based on previous sections, we able to achieved better performance and we didn't need to find optimal hyperparameter for $C_{max}$. For our scorer module, we used MLP scorer with 256 hidden units.

In the decoding phase, we used beam search strategy with beam size 3 to generate the phonemes given the character sequences. For comparison, we evaluated our model with standard global attention and \textit{local-m} attention model \cite{luong2015effective} as the baseline.

\subsection{Result Discussion}
Table \ref{tbl:cmudict} summarizes our experiment on proposed local attention models. We compared our proposed models with several baselines from other algorithm as well. Our model significantly improving the PER and WER compared to encoder-decoder, attention-based global softmax and \textit{local-m} attention (fixed-step size). Compared to Bi-LSTM model which was trained with explicit alignment, we achieve slightly better PER and WER with larger window size ($2\sigma = 3$).

\begin{table}[h]
	\centering
	\caption{Results from baseline and proposed method on G2P task with CMUDict test set}
	\vspace{0.2cm}
	\label{tbl:cmudict}
	\begin{tabular}{|l|c|c|}
		\hline
		\multicolumn{1}{|c|}{\textbf{Model}}                                                     & \begin{tabular}[c]{@{}c@{}}\textbf{PER} \\ \textbf{(\%)}\end{tabular} & \begin{tabular}[c]{@{}c@{}}\textbf{WER} \\ \textbf{(\%)}\end{tabular} \\ \hline
		\multicolumn{3}{|c|}{Baseline}                                                                                                   \\ \hline
		\begin{tabular}[c]{@{}l@{}}Enc-Dec LSTM (2 lyr) \\ \cite{yao2015sequence}\end{tabular} & 7.63              & 28.61             \\ \hline
		\begin{tabular}[c]{@{}l@{}}Bi-LSTM (3 lyr) \\ \cite{yao2015sequence}\end{tabular}      & 5.45              & 23.55             \\ \hline
		\begin{tabular}[c]{@{}l@{}}Att Enc-Dec with \\ Global MLP Scorer (ours)\end{tabular}            & 5.96              & 25.55             \\ \hline
		\begin{tabular}[c]{@{}l@{}}Att Enc-Dec with \textit{local-m} \\ (ours) \cite{luong2015effective}  \end{tabular}            & 5.64              & 24.32             \\ \hline
		\multicolumn{3}{|c|}{Proposed}                                                                                                   \\ \hline
		\begin{tabular}[c]{@{}l@{}}Att Enc-Dec + Unconst (exp)\\ (2$\sigma$ = 2)\end{tabular}     & {5.45}     & \textbf{23.15}             \\ \hline
		\begin{tabular}[c]{@{}l@{}}Att Enc-Dec + Unconst (exp)\\ (2$\sigma$ = 3)\end{tabular}     & \textbf{5.43}              & {23.19}    \\ \hline
	\end{tabular}
\end{table}
\vspace{0.2cm}

\section{Experiment on Machine Translation}
We also conducted experiment on machine translation task, specifically between two languages with similar sentences structure. By using our proposed method, we able to focus only to a small related segment on the source side and the target generation process usually follows the source sentence structure without many reordering process.

\subsection{Dataset}
We used BTEC dataset \cite{kikui2003creating} and chose English-to-France and Indonesian-to-English parallel corpus. From BTEC dataset, we extracted 162318 sentences for training and 510 sentences for test data. Because there are no default development set, we randomly sampled 1000 sentences from training data for validation set. For all language pairs, we preprocessed our dataset using Moses \cite{koehn2007moses} tokenizer. For training, we replaced any word that appear less then twice with unknown (unk) symbol. In details, we keep 10105 words for French corpus, 8265 words for English corpus and 9577 words for Indonesian corpus. We only used sentence pairs where the source is no longer than 60 words in training phase.

\subsection{Model Architecture}
On both encoder and decoder sides, the input words were projected into 256 dims using embedding matrix. We used three Bi-LSTM for our encoder with 512 hidden units for each LSTM (total 1024 hidden unit for Bi-LSTM). For our decoder, we used three LSTM with 512 hidden units. For local monotonic model, we used an MLP with 256 hidden units to generate $\Delta p_t$ and $\lambda_t$. Same as previous section, we only used the unconstrained formulation for local monotonic experiment. For our scorer module, we used MLP scorer with 256 hidden units. In the decoding phase, we used beam search strategy with beam size 5 and normalized length penalty with $\alpha=1$ \cite{wu2016google}. For comparison, we evaluate our model with standard global attention and \textit{local-m} attention model \cite{luo2016learning} as the baseline.

\begin{table}[h]
	\centering
	\caption{Results from baseline and proposed method on English-to-France and Indonesian-to-English translation tasks.}
	\vspace{0.2cm}
	\label{tbl:mt}
	\begin{tabular}{|c|l|}
		\hline
		\textbf{Model}                                                                                             & \multicolumn{1}{c|}{\textbf{BLEU}} \\ \hline
		\multicolumn{2}{|c|}{\textbf{BTEC English to France}}                                                                                           \\ \hline
		\multicolumn{2}{|c|}{Baseline}                                                                                                                  \\ \hline
		\multicolumn{1}{|l|}{\begin{tabular}[c]{@{}l@{}}Att Enc-Dec with \\ Global MLP Scorer\end{tabular}}        & \multicolumn{1}{c|}{49.0}          \\ \hline
		\multicolumn{1}{|l|}{\begin{tabular}[c]{@{}l@{}}Att Enc-Dec with \textit{local-m} \\ (ours) \cite{luong2015effective} \end{tabular}}        & \multicolumn{1}{c|}{50.4}          \\ \hline
		\multicolumn{2}{|c|}{Proposed}                                                                                                                  \\ \hline
		\multicolumn{1}{|l|}{\begin{tabular}[c]{@{}l@{}}Att Enc-Dec + Unconst (exp)\\ (2$\sigma$ = 4)\end{tabular}} & \multicolumn{1}{c|}{\textbf{51.2}} \\ \hline
		\multicolumn{1}{|l|}{\begin{tabular}[c]{@{}l@{}}Att Enc-Dec + Unconst (exp)\\ (2$\sigma$ = 6)\end{tabular}} & \multicolumn{1}{c|}{51.1}          \\ \hline
		\multicolumn{2}{|c|}{\textbf{BTEC Indonesian to English}}                                                                                       \\ \hline
		\multicolumn{2}{|c|}{Baseline}                                       \\ \hline
		\multicolumn{1}{|l|}{\begin{tabular}[c]{@{}l@{}}Att Enc-Dec \\ with Global MLP Scorer\end{tabular}}        & \multicolumn{1}{c|}{38.2}                               \\ \hline
		\multicolumn{1}{|l|}{\begin{tabular}[c]{@{}l@{}}Att Enc-Dec with \textit{local-m} \\ (ours) \cite{luong2015effective} \end{tabular}}        & \multicolumn{1}{c|}{39.8}          \\ \hline
		\multicolumn{2}{|c|}{Proposed}                                       \\ \hline
		\multicolumn{1}{|c|}{\begin{tabular}[c]{@{}l@{}}Att Enc-Dec + Unconst (exp)\\ (2$\sigma$ = 4)\end{tabular}} & \multicolumn{1}{c|}{40.9}                               \\ \hline
		\multicolumn{1}{|c|}{\begin{tabular}[c]{@{}l@{}}Att Enc-Dec + Unconst (exp)\\ (2$\sigma$ = 6)\end{tabular}} & \multicolumn{1}{c|}{\textbf{41.8}}                      \\ \hline
	\end{tabular}
\end{table}

\subsection{Result Discussion}
Table \ref{tbl:mt} summarizes our experiment on proposed local attention models compared to baseline global attention model and \textit{local-m} attention model \cite{luong2015effective}. Generally, local monotonic attention had better result compared to global attention on both English-to-France and Indonesian-to-English translation task. Our proposed model were able to improve the BLEU up to 2.2 points on English-to-France and 3.6 points on Indonesian-to-English translation task compared to standard global attention. Compared to \textit{local-m} attention with fixed step size, our proposed model able to improve the performance up to 0.8 BLEU on English-to-France and 2.0 BLEU on Indonesian-to-English translation task.

\section{Related Work}

Humans do not generally process all of the information that they encounter at once. Selective attention, which is a critical property in human perception, allows attention to be focused on particular information while filtering out a range of other information. The biological structure of the eye and the eye movement mechanism is one part of visual selective attention that provides the ability to focus attention selectively on parts of the visual space to acquire information when and where it is needed \cite{rensink2000dynamic}. In the case of the cocktail party effect, humans can selectively focus their attentive hearing on a single speaker among various conversation and background noise sources \cite{cherry1953some}.

The attention mechanism in deep learning has been studied for many years. But, only recently have attention mechanisms made their way into the sequence-to-sequence deep learning architectures that were proposed to solve machine translation tasks. Such mechanisms provide a model with the ability to jointly align and translate \cite{bahdanau2014neural}. With the attention-based model, the encoder-decoder model significantly improved the performance on machine translation \cite{bahdanau2014neural, luong2015effective} and has successfully been applied to ASR tasks \cite{chorowski2014end, chan2016listen}.

However, as we mentioned earlier, most of those attention mechanism are based on ``global" property, where the attention module tries to match the current hidden states with all the states from the encoder sides. This approach is inefficient and computationally expensive on longer source sequences. A ``local attention" was recently introduced by \cite{luong2015effective} which provided the capability to only focus small subset of the encoder sides. They also proposed monotonic attention but limited to fixed step-size and not suitable for a task where the length ratio between source and target sequence is vastly different. Our proposed method are able to elevated this problem by predicting the step size dynamically instead of using fixed step size. After we constructed our proposed framework, we found work by \cite{raffel2017online} recently that also proposed a method for producing monotonic alignment by using Bernoulli random variable to control when the alignment should stop and generate output. However, it cannot attend the source sequence outside the range between previous and current position. In contrast with our approach, we are able to control how large the area we want to attend based on the window size. 

\cite{chorowski2014end} also proposed a soft constraint to encourage monotonicity by invoking a penalty based on the current alignment and previous alignments. However, the methods still did not guarantee a monotonicity movement of the attention.

To the best of our knowledge, only few studies have explored about local and monotonicity properties on an attention-based model. This work presents a novel attention module with locality and monotonicity properties. Our proposed mechanism strictly enforces monotonicity and locality properties in their alignment by explicitly modeling them in mathematical equations. The observation on our proposed model can also possibly act as regularizer by only observed a subset of encoder states. Here, we also explore various ways to control both properties and evaluate the impact of each variations on our proposed model. Experimental results also demonstrate that the proposed encoder-decoder model with local monotonic attention could provide a better performances in comparison with the standard global attention architecture and \textit{local-m} attention model \cite{luong2015effective}.

\section{Conclusion}
This paper demonstrated a novel attention mechanism for encoder decoder model that ensures monotonicity and locality properties. We explored various ways to control these properties, including dynamic monotonicity-based position prediction and locality-based alignment generation. The results reveal our proposed encoder-decoder model with local monotonic attention significantly improved the performance on three different tasks and able to reduced the computational complexity more than one that used standard global attention architecture.

\section{Acknowledgement}
Part of this work was supported by JSPS KAKENHI Grant Numbers JP17H06101 and JP17K00237.

\bibliography{ijcnlp2017}
\bibliographystyle{ijcnlp2017}

\end{document}